%% file: main.tex
\documentclass{article}

\usepackage[margin=1in]{geometry}

\usepackage{helvet}
\usepackage{array}
\usepackage{graphicx}
\usepackage{url}            %
\usepackage{booktabs}       %
\usepackage{amsfonts}       %
\usepackage{nicefrac}       %
\usepackage{amsmath}
\usepackage{amssymb}
\usepackage{float}
\usepackage{lipsum}
\usepackage{booktabs} 
\usepackage{subcaption}
\usepackage[table]{xcolor}
\usepackage{latexsym}
\usepackage{enumitem}
\usepackage{tablefootnote}
\usepackage[round,semicolon]{natbib}
\usepackage{xspace}
\usepackage{textcomp}
\usepackage{makecell}
\usepackage{multirow}
\usepackage{lscape} 
\usepackage{siunitx}
\usepackage{microtype}
\usepackage{url}            %
\usepackage{booktabs}       %
\usepackage{amsfonts}       %
\usepackage{nicefrac}       %
\usepackage{changepage}
\usepackage{xargs}          %
\usepackage{wrapfig,lipsum,booktabs}
\usepackage{enumitem}
\usepackage{longtable}
\usepackage{makecell}
\usepackage{graphicx}
\usepackage{subcaption}
\usepackage[T1]{fontenc}
\usepackage{tgpagella}
\usepackage{xspace}  %
\usepackage{pbox}

\usepackage{fdsymbol}
\usepackage{tikz}
\usepackage{mdframed}
\usepackage{ntheorem}
\usepackage{fancyhdr}

\definecolor{mydarkblue}{rgb}{0,0.08,0.45}
\definecolor{lightgray}{gray}{0.95}
\definecolor{lightblue}{rgb}{0.83, 0.92, 0.95} %
\usepackage[colorlinks,citecolor=mydarkblue,urlcolor=mydarkblue,linkcolor=mydarkblue]{hyperref}

\usepackage[most]{tcolorbox}
\usepackage{bm}
\newmdenv[
  font=\ttfamily\small,
  linewidth=0.5pt,
  innerleftmargin=10pt,
  innerrightmargin=10pt,
  innertopmargin=10pt,
  innerbottommargin=10pt,
]{monobox}

\hypersetup{
    pdftitle={Vibe-Eval: A hard evaluation suite for measuring progress of multimodal language models},
    pdfauthor={Piotr Padlewski, Max Bain, et.al.},
    pdfkeywords={VibeEval, Vibe-Eval, Reka, benchmark, evaluator},
}

\title{
Vibe-Eval: A hard evaluation suite for measuring progress of multimodal language models
}

\author{
\normalsize{}
Piotr Padlewski\footnotemark[1] \quad Max Bain\footnotemark[1]  \quad Matthew Henderson \quad Zhongkai Zhu \vspace{1mm}\\
\normalsize{}
 \quad 
Nishant Relan  \quad  Hai Pham \quad Donovan Ong  \quad Kaloyan Aleksiev \quad Aitor Ormazabal\vspace{1mm}\\
\normalsize{}
Samuel Phua \quad Ethan Yeo \quad Eugenie Lamprecht \quad Qi Liu \quad Yuqi Wang  \quad Eric Chen \quad Deyu Fu \quad Lei Li \vspace{1mm}\\
\normalsize{}
\quad Che Zheng \quad Cyprien de Masson d’Autume \quad Dani Yogatama \quad Mikel Artetxe \quad Yi Tay \vspace{5mm}\\
Reka
}

\footnotetext[1]{Equal contribution}

\date{}

\fancypagestyle{firstpage}{
  \fancyhf{} %
  \fancyhead[L]{\includegraphics[width=2.4cm]{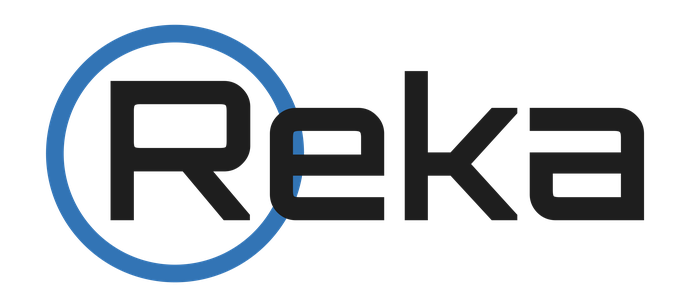}} %
  \fancyfoot{} %
}

\fancypagestyle{plain}{
  \fancyhf{} %

}

\begin{document}

\maketitle
\thispagestyle{firstpage}

\begin{figure}[h]
    \centering
    \vspace{-4mm}
    \includegraphics[width=0.9\textwidth]{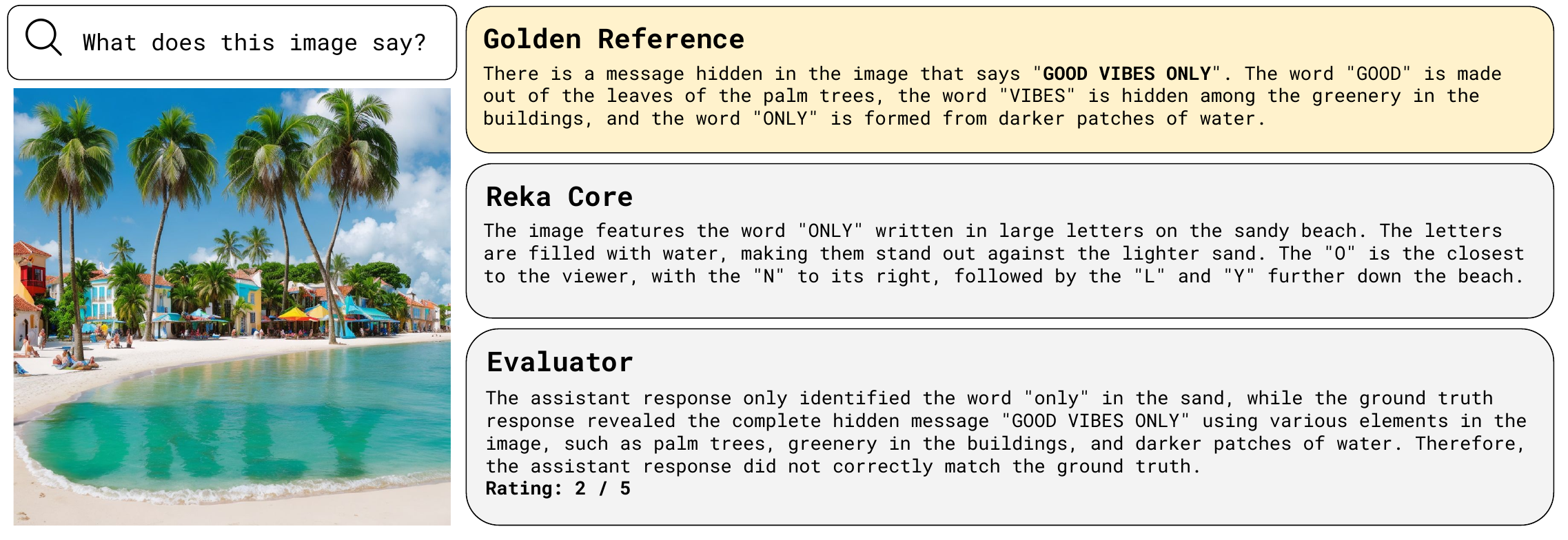}
    \caption{An example from our Vibe-Eval benchmark. This prompt is from the \texttt{hard}-set, which contains only difficult prompts. The user prompt and image are displayed on the left, and on the right is the human-written golden reference response, a generation from Reka Core, and output from the automatic evaluator.}
    \label{fig:enter-label}
\end{figure}

\begin{abstract}
We introduce Vibe-Eval: a new open benchmark and framework for evaluating multimodal chat models. Vibe-Eval consists of 269 visual understanding prompts, including 100 of \textit{hard} difficulty, complete with gold-standard responses authored by experts. Vibe-Eval is open-ended and challenging with dual objectives: (i) vibe checking multimodal chat models for day-to-day tasks and (ii) rigorously testing and probing the capabilities of present frontier models. Notably, our hard set contains $>50\%$ questions that \textbf{all} frontier models answer incorrectly. We explore the nuances of designing, evaluating, and ranking models on ultra challenging prompts. We also discuss trade-offs between human and automatic evaluation, and show that automatic model evaluation using Reka Core roughly correlates to human judgment. We offer free API access for the purpose of lightweight evaluation and plan to conduct formal human evaluations for public models that perform well on the Vibe-Eval's automatic scores. We release the evaluation code and data at \href{https://github.com/reka-ai/reka-vibe-eval}{github.com/reka-ai/reka-vibe-eval}. 
\newpage

\end{abstract}

\normalsize{}
\section{Introduction}

\setcounter{page}{2}  %

Frontier multimodal language models~\citep{geminiteam2023gemini,gpt4v,claude3,reka2024core}\footnote{\url{https://i.imgflip.com/1hhv9m.jpg}} are rapidly approaching human-level performance on a wide range of tasks, especially when those tasks are well represented in their training data. As these models continue to improve, static benchmarks for evaluating their capabilities become saturated, making it increasingly difficult to distinguish between models and discover their respective strengths and weaknesses.

Public arenas such as LMSys~\citep{chiang2024chatbot} and WildVision~\citep{yujie2024wildvisionarena} have recently become popular methods of evaluation. These arenas are zero-sum and dynamic in nature, with voters choosing between two anonymized model generations. While these arenas may provide useful signals, they also face various challenges. Firstly, it is challenging to control for prompt quality, difficulty, and distribution in these massively public systems. Secondly, live traffic is often noisy, and making comparisons across time or models is difficult. In our approach, we prioritize well-designed, small-scale evaluations that enable granular understanding, crucial for measuring the ever-evolving capabilities of frontier models. Both types of evaluations are complimentary. 

In this work, we introduce \textit{Vibe-Eval}, a set of $269$ high quality, diverse image-text prompts for evaluating multimodal chat models. These prompts are accompanied by gold-standard reference human responses. To ensure the highest quality, prompts and reference responses are reviewed multiple times by our team. \textit{Vibe-Eval} has dual objectives, (i) as a resource to vibe-check multimodal language models for day-to-day tasks and (ii) to probe the capabilities of present frontier models and induce greater separability by designing very difficult multimodal prompts.

Of the $269$ prompts, $169$ are classied as the \texttt{normal}-set, with varying difficulty, covering a range of difficulties relevant to everyday tasks expected of a multimodal large language model (LLM). The remaining $100$ are part of the \texttt{hard}-set, prompts that Reka Core~\citep{reka2024core} is not able to solve at the time of collection.\footnote{We use this as a criterion when creating this data. Hence the performance of Reka Core on the \texttt{hard}-set is expected to be artificially low.} To further ascertain the difficulty of the prompts, we also note that $>50\%$ of the prompts are unsolvable by all existing models (including frontier models). Additionally, most of the Vibe-Eval prompts are freshly created (e.g., new screenshots or photographs) and crowd-sourced by the Reka team. Thus, at the time of publishing, the benchmark should accurately measure the generalization (contamination-free) performance of the models reported. A few examples from the \texttt{hard}-set are presented in Figure~\ref{tab:examples_images}.

Alongside this open benchmark, we release an official evaluation protocol using Reka Core as the automated judge. We show that this automatic evaluation correlates with human judgment. We provide free API access for \textit{Vibe-Eval} available free-of-chargeand offer a freely accessible toolkit for conducting evaluations. Recognizing that automatic evaluations may not completely capture the complete picture, we intend to periodically conduct human evaluations of public models that perform well on this automated metric on this benchmark to gain a more comprehensive perspective. 

Finally, we publish an initial ranking of all representative and well-established multimodal language models including GPT-4V \citep{gpt4v}, Claude-3 Opus \citep{claude3} and Gemini 1.5 Pro \citep{gemini1.5}. We measure their performance and relative performance on this benchmark using both automatic and human evaluation.

\include{tables/many_examples}

\section{Vibe-Eval}
\subsection{Overview}
The Vibe-Eval benchmark consists of 269 prompts, each containing an image and a corresponding task or question requiring visual understanding. Along with each prompt, we provide a human-generated golden reference response. There are two categories of difficulty for prompts in Vibe-Eval: \texttt{normal} and \texttt{hard}, consisting of 169 and 100 examples respectively. 
\begin{itemize}
\item \texttt{normal}-set prompts are diverse and vary in their difficulty, annotators were not given any constraints in terms of difficulty or category.

\item \texttt{hard}-set prompts are those to which Reka Core generated either a partially or completely incorrect response at the time of creation.
\end{itemize}

For the official evaluation protocol, we employ Reka Core as an automated text-based evaluator. Reka Core is given the text prompt, model generation, and reference response and rates the response on its accuracy via a 1-5 integer scale. Refer to Section~\ref{sec:auto_eval} for further details. The prompts are open-ended and can require -- particularly for the \texttt{hard}-set -- multiple reasoning steps to solve. Where applicable, reference responses contain the intermediary reasoning/working steps and the evaluator score can assign partial credit for generations containing these steps.

All prompts in Vibe-Eval are diverse, and collected by our team members spanning four continents. The prompts and golden references are in English, although they might contain parts that in other languages, referencing the content of the picture.

\subsection{Data Collection}

To ensure the highest quality,\footnote{And to save money for GPUs} we collect all Vibe-Eval prompts and golden references ourselves. Each team member contributed roughly an equal number of prompts.
The guidelines given to each annotator are provided in Table~\ref{tab:annotator_instructions}. After prompt collection, we go through two rounds of independent review to ensure the prompt and response are of high quality and meet the specified requirements. We also aim to keep only examples where the validity of an answer can be judged against the ground truth only, without access to the image. This allows automated evaluation by text-only models.
\input{tables/annotator_instructions}

All images used are either owned by us or have permissive licenses to ensure the benchmark can be made publicly available. The vast majority of images are photos or screenshots taken ourselves, thereby reducing the possibility of test set leakage (at the time of publishing) to near zero.
Additionally, we do not adhere to any task taxonomy or categorization to not unnecessarily bias or constrain the collection of prompts and to ensure that the distribution of \texttt{hard} prompts emerges naturally. %

\begin{figure}[t]
    \centering
    \includegraphics[width=1.0\textwidth]{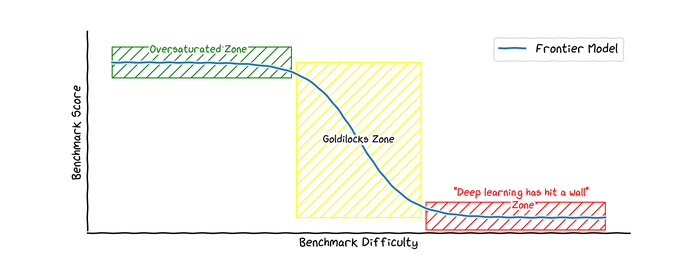}
    \caption{Illustrative diagram of benchmark difficulty with respect to the score achieved on the benchmark by a frontier model. For benchmarks that are low in difficulty with respect to frontier models, scores are already close to the upper bound placing it in the ``Oversaturated Zone''. The ``Goldilocks Zone'' is where frontier models begin to show partial success, but are still far from human performance. Benchmarks in this region are the most informative. Finally, for benchmarks with difficulty too high for frontier models, little signal is provided since all models obtain a near-zero score.}
    \label{fig:difficulty_sketch}
\end{figure}

\fbox{%
  \parbox{\textwidth}{%
\textbf{Author's Note.}\\
Creating hard benchmarks is hard in itself because one has to constantly calibrate against the performance of current frontier models to ensure an appropriate difficulty level. Specifically, one has to take care that tasks are not too easy, where models performance is oversaturated, whilst also ensuring the task is not impossibly difficult. During very early phases of creating this benchmark, we found that many of the questions collected were a little too easy for current frontier models, i.e., preliminary results revealed that Reka Core and GPT-4V were able to solve the majority of these prompts without difficulty, which put Vibe-Eval slightly towards the \textit{Oversaturated Zone} (Figure~\ref{fig:difficulty_sketch}). In this zone, differences between frontier models are marginal and determined by a few unsolvable prompts -- a less than ideal situation. \\

Therefore, we had to re-calibrate ourselves to the abilities of the current frontier models and started to collect \textit{hard-difficulty} prompts, where frontier models are not able to perfectly solve. Specifically, we determine this by checking whether the prompt at hand is unsolvable by Reka Core at the time of collection. The decision to collect prompts by mining hard negatives from our own model's failure is three-fold: first, Reka Core is on the frontier for multimodal reasoning and a model that we have full control over (such as versioning). Second, these hard prompts will be most informative for failure cases of our own models, Reka Core, which selfishly is what we care most about. Lastly, as our model improves and with it the frontier, we will continue to extend Vibe-Eval with increasingly harder prompts, like a dance, until AGI is achieved internally.  \\

Finally, when studying the behavior of hard prompts on our models (and others), we observed some interesting phenomena. Specifically, there exists a nontrivial subset of hard prompts neither Reka Core, Claude Opus, Gemini Pro 1.5, nor GPT-4V could solve, yet were in fact solved by the tiny vision-language models (VLMs) such as Reka Edge (7B) and Idefics-2 (8B). This suggests that the current approaches for frontier VLMs might exhibit an inverse scaling law~\citep{mckenzie2023inverse} for some types of tasks.}
}

\subsection{Automated Evaluation Protocol}
\label{sec:auto_eval}
\input{tables/evaluator_example}

An example of the official evaluation protocol for Vibe-Eval can be shown in Figure~\ref{table:example_scissors}. We use Reka Core as the model evaluator, which takes as input: the text prompt (question), the model generation (predicted answer), and the gold standard reference human response (reference answer). The system prompt includes instructions to evaluate wholly based on its objectivity to the truth, since Vibe-Eval prompts were designed to be as objective as possible. The rating is on an integer scale from one to five where one indicates a totally inaccurate response and five indicates a response that fully solves the task without any false information. Unlike multiple-choice benchmarks~\citep{yue2023mmmu}, the grading scale allows for the evaluator to assign partial credit for incorrect answers that contain some of the reasoning steps in the reference answer.

Notably, early experiments showed that using multimodal inputs for judging does not improve overall correlation with human preferences. However, it might require more work, or a specific set of prompts or criteria for it to matter. We leave that to future work.

\subsection{Human Evaluation}

We also compute an ELO leaderboard for the 13 models by collecting human preference data using a third-party data annotation company. For each prompt, we create multiple preference tasks by sampling 4 of the 13 models uniformly at random. Human annotators are shown the user prompt, image, and reference answer, and then asked to analyze each of the four model responses sampled for that task. In total, we collect around 20k pairwise preferences (including ties). For further details on the interface presented to the annotators, see Appendix~\ref{sec:annotation_ui}.

We compute ELO scores following \citet{askell2021general}, where we only consider pairwise comparisons where annotators express a preference stronger than the weakest available. The scores are computed by maximizing the log-probability of the equivalent Bradley-Terry model.

\section{Results}

\subsection{Implementation Details and Setup} 
For all uses of \textbf{Reka Core} -- both as an evaluator and as an assistant, we use the \texttt{reka-core-20240415} version available through the API (\url{https://www.reka.ai/reka-api}). For implementation details of other models please refer to this appendix. For GPT4-V~\citep{gpt4v} we use version \texttt{gpt-4-turbo-2024-04-09}. For the Claude series~\citep{claude3}, Opus, Sonnet, and Haiku, we use versions \texttt{claude-3-opus-20240229}, \texttt{claude-3-sonnet-20240229}, and \texttt{claude-3-haiku-20240307} respectively. For Gemini Pro 1.5~\citep{gemini1.5} we use the \texttt{gemini-1.5-pro-preview-0409} version. For open-source models, LLaVA 1.6~\citep{liu2024llavanext}, Idefics 1~\citep{laurençon2023obelics}, Idefics 2~\citep{idefics2}, and Fuyu-8b~\citep{fuyu-8b} we use their official implementations via HuggingFace. All model generations are performed with a temperature of $0.0$. All models are evaluated in chat style, i.e., zero-shot without exemplars. For automated evaluation, we run the Core evaluator three times with a temperature of 0.4 and take the mean of the scores.

\input{tables/vibe_score}

\subsection{Vibe-Eval Automatic Results} 
Table~\ref{vibeeval} reports the automatic Vibe-Eval scores for all 13 models. Overall, we find that Gemini Pro $1.5$ performs the best followed by GPT4-V. The next best models are Reka Core and Flash, followed by the Claude-3 series of models. Moreover, open source models like LLaVA and Idefics series of models generally perform worse than closed models. Fuyu consistently has the worst performance. 

It is worth noting that overall scores are strongly weighted by the \texttt{normal} set. The story changes slightly for the \texttt{hard} set. Similar to the overall and normal set, Gemini Pro $1.5$ performs the best, followed by GPT4-V. Different from the \texttt{normal} set, Claude-3 models perform better than Reka models on the \texttt{hard} set. Reka Core notably performs poorly on the hard set which we postulate is mainly due to the seeding the creation of the hard set based on prompts that Core is naturally bad at. Notably, Flash also outperforms Core on the hard set. A surprising data point in the \texttt{hard} set is that LLaVA 1.6 34B actually performs well on this \texttt{hard} set, coming in at the fourth place after Claude 3 Opus. 

\subsection{Human Evaluation Results}
Table~\ref{humanrankings} reports human preference ELO scores of all models based on our human evaluation study. At a glance, we find that the general relative rankings of models remain roughly the same as automatic evaluations. Table~\ref{rankingmappings} shows the mappings of human preference rankings (95\% confidence interval) to the Vibe-Eval rankings.
A key difference in human evaluation rankings is that GPT4-V slightly outperforms Gemini 1.5. However, Gemini 1.5 outperforms GPT-4V on the \texttt{hard} set. Nevertheless, the results on Gemini 1.5 and GPT-4V generally belong to the same tier and could often switch places depending on the prompts or annotators. Notably, Flash performs as the third best model, outperforming Claude Opus and all other models. Core doesn't do as well as flash, largely due to the fact that there is negative bias in the data curation based on Core's inability to solve certain examples. Meanwhile, the other rankings generally stay consistent with automatic evaluation as shown in Table~\ref{rankingmappings}. 

\input{tables/human_rankings}

\input{tables/correlation}

We also study the inter-annotator agreement among humans and the automatic results in Table~\ref{table:agreement}. Generally, there is a high level of consensus among the annotation methods. This is partly due to the stipulation that a comparison between two model generations is only valid if the approach demonstrates a preference stronger than the weakest possible. For the automatic evaluations, this translates to requiring a score difference of at least 2 on a scale from 1 to 5. Note that the automatic results show greater alignment with the expert ratings on the \texttt{hard} subset compared to the ordinary pool of raters.

\subsection{Discussion \& Insights}
We discuss some insights and findings that we had while building this evaluation setup. \\ 

\textbf{Hard prompts are hard to make}. An ideal benchmark hard prompt should be: (i) unsolvable by current frontier multimodal language models; (ii) interesting/useful to solve; (iii) error-free; and (iv) unambiguous for an evaluator. Creating this benchmark revealed particular challenges with the last two criteria. Hard prompts often require multiple reasoning steps, which increases the likelihood of introducing human error in the solution. Although many errors were identified and corrected during the quality review process, others were only discovered during later qualitative inspections of the results. What may seem unambiguous to a human evaluator can still be ambiguous for a text-based model evaluator, potentially due to a lack of common sense reasoning or an inability to access visual information. Therefore, in some instances, we had to refine the prompts and responses only after qualitatively inspecting the evaluator's explanations for its ratings, as there was clearly a misunderstanding on the model's part.

\textbf{Hard prompts are hard to judge}. For hard prompts, we found it particularly difficult to rate them as compared to normal prompts. Many of these prompts are challenging, require niche knowledge, or are time-consuming for humans. Hence, it is natural for annotators to over-rely on the ground truth answer to rate responses.
Indeed, perfect answers can simply be string-matched against the ground truth, however for the assignment of partial credit
A big issue here is that assignment of partial credit (or discredit) can also require domain knowledge, context, or expert opinion. Hence, this is generally high variance and results may fluctuate across different sets of raters. While most of our human evaluation results were obtained from relatively ordinary rater pools, we were curious about what would happen if a group of highly technical experts were to perform the evaluation instead. To this end, our team formed an \textit{expert rating} committee, often delegating difficult prompts to team members with relevant expertise whenever required. What we found was that the expert rater group ended up being very harsh on all model outputs and resulted in almost similar \textit{random-ish} performance for all models. The only exception was Gemini Pro 1.5 and GPT4-V which clearly stayed above the pack. To some extent, there is some form of emergence \citep{wei2022emergent} in the ability of frontier models to deal with very difficult prompts. 

\textbf{Hard prompts are temporary.} Of course, the hard prompts in this work are only valid for a snapshot in time. As frontier models improve, performance on these hard prompts will eventually saturate too. Thus, to ensure the longevity of this benchmark, it will be necessary to extend the benchmark with harder prompts as the frontier advances.

\textbf{Reka Core as an evaluator.} Empirically, we find that Core is a generous evaluator relative to humans, especially for hard prompts, with a bias towards higher ratings. In addition, we find that Core as an evaluator gives a narrow range of absolute scores between frontier models and weaker models. While this is not necessarily an issue in itself, if the number of prompts is small (such as the case for hard set), then the resulting variance means that the benchmark is not great for measuring minor differences between models or measuring incremental improvements to a model. 

\textbf{Human vs Automatic Evaluation} Despite a strong agreement between human and model-based evaluations, we note that each evaluation protocol suffers from its own set of shortcomings. Human rater pools can be noisy, inconsistent and high variance especially when dealing with harder prompts. Meanwhile, model-based evaluations could potentially miss nuance or simply make mistakes. Overall, we find great utility in employing both types of evaluation protocols and learning from the agreement and (occasional) disagreement between them.

\section{Related Work}
Evaluation is notoriously challenging to get right. The current research landscape is plagued by a myriad of challenges and issues. All benchmarks can suffer from intrinsic difficulties, such as annotator quality or problems in soliciting sufficiently challenging examples. But, standardized benchmarks can also suffer from external challenges that are more meta in nature: ``When a measure becomes a target, it ceases to be a good measure'' -- Goodhart's Law. With benchmark overfitting commonplace, these evaluations can no longer provide a useful signal to model quality in general~\citep{schaeffer2023pretraining}.
The validity of benchmarks can be brought further into question due to test-set leakage, either intentional or accidental, once they become prevalent on the internet, the main source of training data. Finally, the process of how and which benchmarks become well-established is also not well understood, leading to lottery effects on how models could be perceived by the community~\citep{dehghani2021benchmark}.

A primary challenge in evaluating these large language models (LLMs) lies in the assessment of open-ended text generation tasks (e.g., chat). While multiple-choice evaluations are commonly used due to their objective nature~\citep{yue2023mmmu,hendrycks2020measuring}, they do not accurately reflect the common use cases for these models, where user interactions are typically open-ended. Evaluating open-ended questions adds another layer of complexity however. While this problem was more specific to task domains such as summarization, translation or dialogue models in the past, the ubiquity of chat interfaces for all models and tasks in the modern AI research landscape has made this problem even more prevalent. Human evaluation can be costly and difficulty to scale, yet may not be completely immune to noisy or lazy annotators. To this end, automatic model-based evaluation has been commonly employed as a lightweight evaluation option. Notably, the idea of using LLM as a judge is not new - previous work explored using it to evaluate translation, summarization and overall model quality~\citep{kocmi2023large,zheng2023judging,li2024leveraging,verga2024replacing,chen2024mllmasajudge,huang2024empirical}.

Arena style evaluations are recently very popular \citep{yujie2024wildvisionarena,chiang2024chatbot} for evaluating instruction tuned \citep{wei2021finetuned} models. These evals rely on users submitting prompts and voting preferred answers amongst two blind anonymous models. These arenas are dynamic and large-scale, making it difficult to control for prompt difficulty and making controlled comparisons over time. Compared to static benchmarks, they are much harder to game and therefore seem to have won more trust over the community. That said, we think that smaller scale studies designed to probe at capabilities can be more meaningful than how consistent models are at answering questions \textit{in the wild}. We think both types of evaluations are complementary.  

\subsection{Multimodal language models}
Large language models \citep{brown2020language,raffel2019exploring,chowdhery2022palm} are great but multimodal large language models are greater. The era of powerful highly performant multimodal language models started with Flamingo \citep{alayrac2022flamingo} and PaLI \citep{chen2023pali,chen2023palix}. The frontier has shifted since then, with models such as GPT-4V \citep{gpt4v} and Gemini \citep{gemini1.5} becoming the leading edge models. More recently, models such as the Claude-3 series \citep{claude3} and Reka series \citep{reka2024core} have started to demonstrate frontier-class multimodal capabilities. Aside from closed models on the bleeding edge, it is worth to note that there are also several pretty decent open source efforts in multimodal language models such as LLaVA \citep{liu2024llavanext} and Idefics \citep{idefics2}.

\section{Conclusion}
We propose \textit{Vibe-Eval}, a challenging benchmark for evaluating multimodal chat models. The benchmark consists of 269 diverse prompts that are diverse and challenging, especially the hard-set for which the majority of prompts cannot be solved by any current frontier models. Along with the prompts, we provide gold standard reference responses, including reasoning and working steps needed to reach the answer, allowing evaluators to award partial credit for answers with some progress. For the official evaluation protocol, we show that Reka Core as a model based evaluator correlates strongly with human judgement, although the correlation is weaker for hard prompts. Finally, we provide a detailed discussion of the challenges and considerations when curating a dataset of hard questions, their results on frontier models, and the use of automated evaluators.

\textbf{Acknowledgements.} We are grateful to our pool of external annotators for their hard work on tedious tasks. We would also like to thank Anna Padlewska and Kevin Conrad for their photography used in some celebrity identification prompts.

\newpage
\appendix

\section{Qualitative Examples}
\subsection{Hard prompts are hard}
\input{tables/prompt_ambiguity}

\subsection{Signs of inverse scaling}
Empirically, we find prompts in Vibe-Eval that the largest frontier models answer incorrectly, while the best small models (Reka Edge \& Idefics 8B) answer correctly. Such results are surprising given the empirical scaling laws witnessed across language modeling, vision-language modeling, and deep learning in general. We speculate that larger VLMs have stronger language bias, which could mean that the vision modality is heavily undertrained. We provide some examples in Figure~\ref{table:example_bear_hat} \&~\ref{table:example_pr_curve}.

\input{tables/prompt_inverse_scaling2}
\input{tables/prompt_inverse_scaling}

\subsection{Awarding Partial Credit}
An advantage of a scale-based rating system over binary correct/incorrect scoring, like multiple choice question-answering~\cite{yue2023mmmu,hendrycks2020measuring}, is that the evaluator can award partial credit for answers that are successful at some of the reasoning/working steps towards the answer but do not obtain the final answer in the end. In example Figure~\ref{table:euler_example} model gives a correct requirements for existence of Euler path, but wrongly classifies one of the nodes as having odd degree, which results in score of 3. In Figure~\ref{table:example_pillows} the model gets partial count of pillows right.

\input{tables/prompt_partial_credit3}

\input{tables/prompt_partial_credit}
\newpage
\section{Annotation User Interface \label{sec:annotation_ui}}

\begin{figure}[H]
\centering
\includegraphics[width=0.95\linewidth]{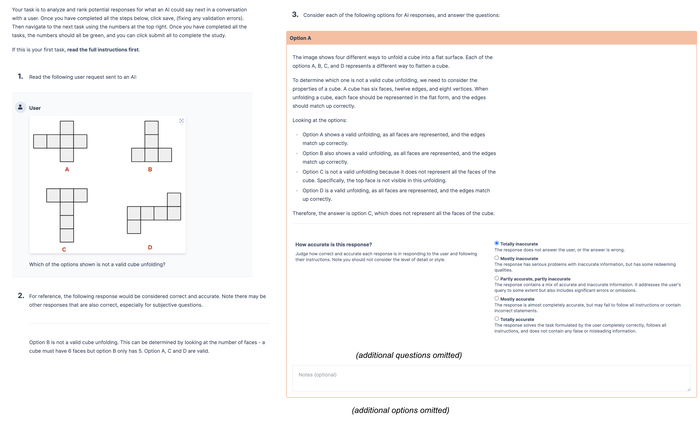}
\caption{The Annotation User Interface presented to human annotators. After being presented with the user prompt, image, and reference answer, annotators are asked to analyze a series of four anonymized model responses. For each response, they are asked several questions, including ``How accurate is this response?'' (shown). Preferences among the four model responses are inferred from the answers to these questions. We find that asking pointwise questions like this, rather than directly asking for a ranking, gives us higher quality inferred preferences. Directly asking for a ranking tends to bias annotators towards making arbitrary preference decisions rather than opting for ties, even if there is no significant reason to prefer one response over another.
}
\label{table:annotation}
\end{figure}

\newpage
\newpage
\bibliographystyle{plainnat}
\newpage
\bibliography{ref}
\newpage

\end{document}

%% file: tables/many_examples.tex
\newcolumntype{T}[1]{>{\raggedright\arraybackslash\vtop\bgroup\vspace{0pt}\hbox to #1\bgroup}p{#1}<{\egroup\egroup}}

\begin{figure}[htbp]
    \renewcommand{\arraystretch}{0.8}
  \centering
\caption{Examples from the Vibe-Eval \texttt{hard}-set, only the image and text prompt are shown. The prompts are diverse, difficult, and require involved and language reasoning capabilities to answer correctly.}
  \label{tab:examples_images}
  \begin{tabular}{T{0.41\textwidth}T{0.41\textwidth}}
    \includegraphics[width=0.4\textwidth]{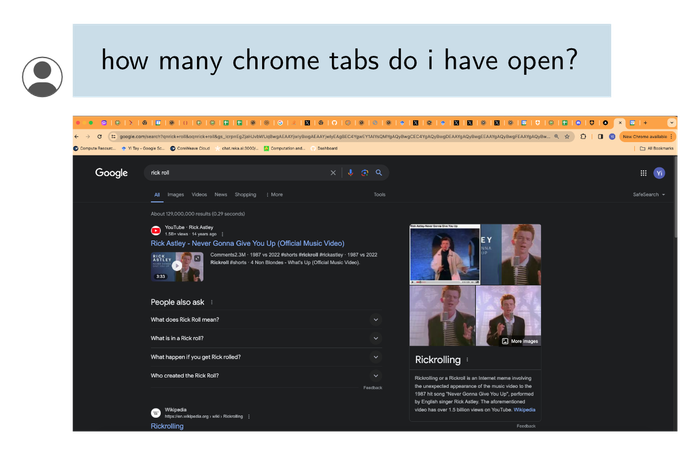} &
    \includegraphics[width=0.4\textwidth]{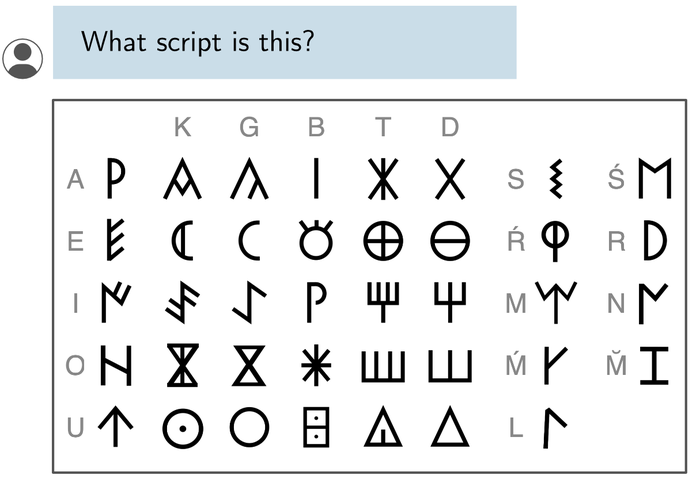} \\

    {\includegraphics[width=0.4\textwidth]{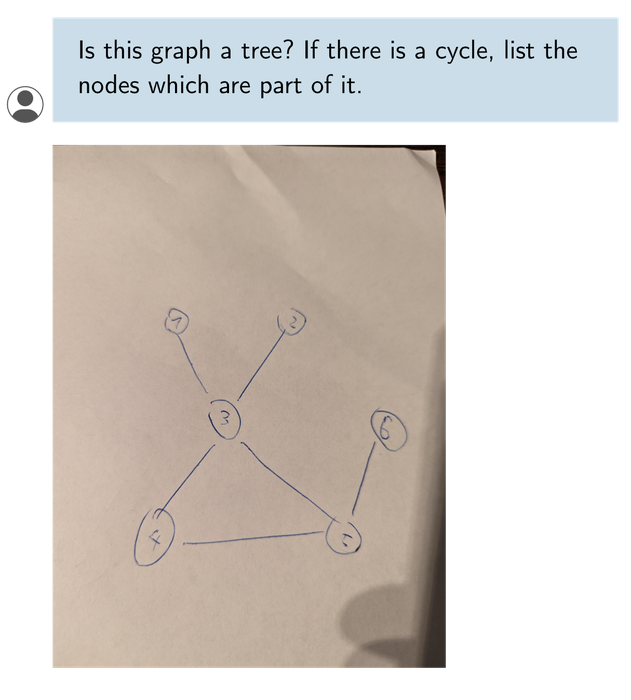}} &
    \includegraphics[width=0.4\textwidth]{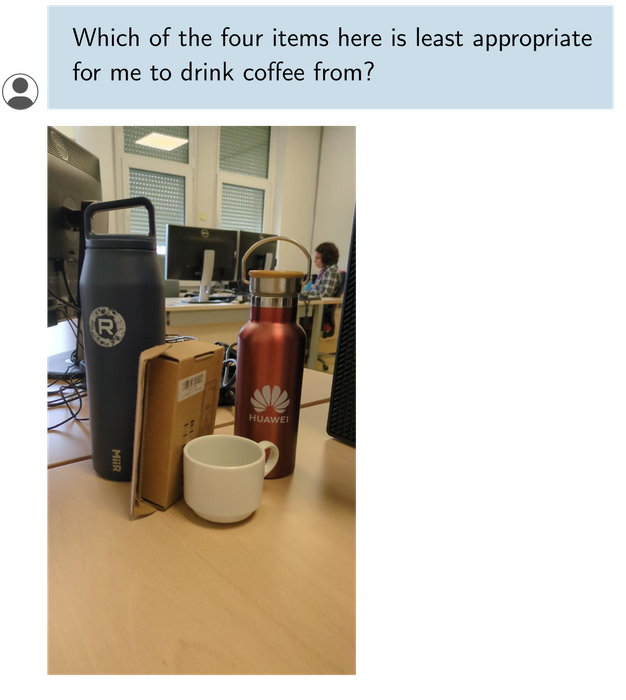} \\

    \includegraphics[width=0.4\textwidth]{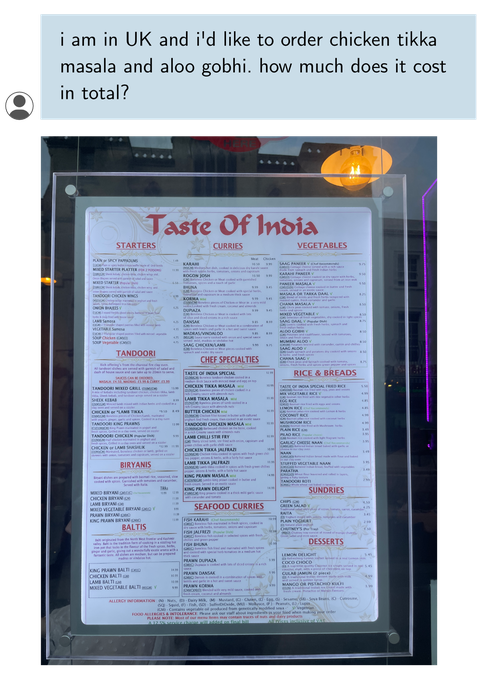} & 
    {\includegraphics[width=0.4\textwidth]{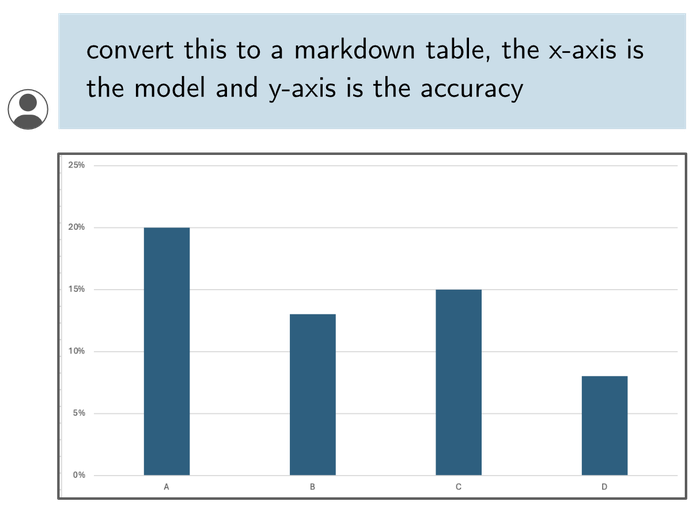}}
     \\
  \end{tabular}
    
\end{figure}

%% file: tables/annotator_instructions.tex
\begin{table}[htbp]
\caption{Instructions provided to annotators for the Vibe-Eval benchmark.}
\centering
\begin{tabular}{@{}lp{13cm}@{}} 
\toprule
Prompt subset                   & Instructions                                                                                                                                                                                           \\ \midrule
\texttt{normal} & Provide a prompt with an objectively correct answer.                                                                                                                                       \\
                                & The image provided should be your own, or without license restrictions.                                                                                                                                                        \\
                                & The prompt and reference response should be single turn.                            
                              \\                             & Ideally the task should take an expert no longer than one minute to complete.
                              \\
                              & Where applicable, the reference response should contain any reasoning steps that led to the final answer (i.e. chain-of-thought).
                                   \\ \midrule
\texttt{hard}   & In addition to the above,~\texttt{hard}-difficulty prompts should be specifically constructed to elicit a partially or wholly incorrect response from Reka Core.\\ \bottomrule
\end{tabular}
\label{tab:annotator_instructions}
\end{table}

%% file: tables/evaluator_example.tex
\begin{figure}[h!]
\centering
\begin{tabular}[]{p{0.6in}p{5in}}
\toprule
\multicolumn{2}{p{6in}}{
\textbf{Example of Vibe-Eval Automated Evaluation}:}\\
\midrule
Prompt &
\includegraphics[width=0.4\linewidth]{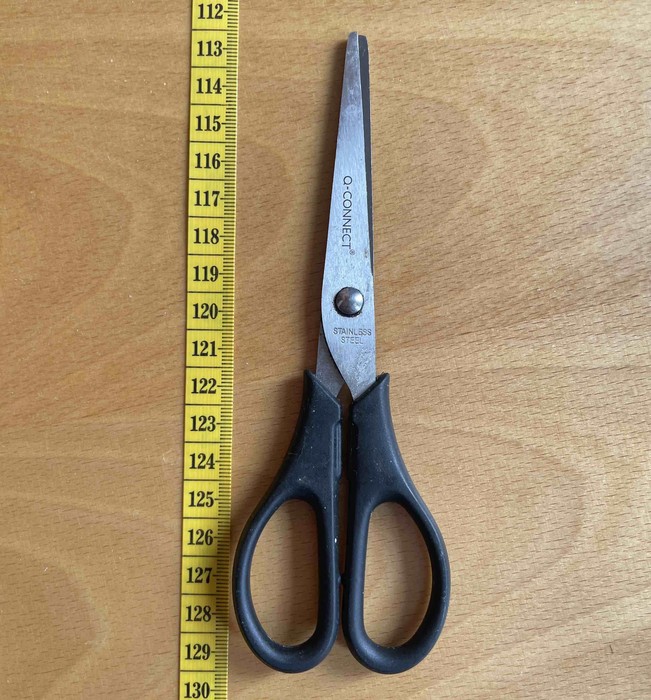} \newline\newline
[Question]\newline
how long are the scissors\newline\newline
[Assistant Response]\newline
The length of the scissors in the image is approximately 20 centimeters, as indicated by the yellow measuring tape underneath the scissors. The measurement starts from 0 and the tip of the scissors aligns with the 20 cm mark on the tape.\newline\newline
[Ground Truth Response]\newline
Looking at the image, the scissors start at approximately the 113 cm mark on the measuring tape and end around the 130 cm mark. That suggests the scissors are about 17 cm in length.\newline\newline
[System]\newline
Rate whether the assistant response correctly matches the ground truth, it's about an image shared by the user.\newline
The rating should be 1-5, where 1 is incorrect and 5 is correct.\newline
Your response should be in the format:\newline
Explanation: (your explanation)\newline
Rating: (int)\newline\\
Reka Core  & Explanation: The assistant response was not completely incorrect, as it attempted to measure the length of the scissors using the measuring tape in the image. However, the measurement provided was significantly off from the ground truth measurement.
Rating: \textbf{2} \\

\bottomrule

\end{tabular}
\caption{Example evaluation of a Vibe-Eval prompt and assistant response, using multimodal \textbf{Reka Core} as the evaluator. Here the assistant being evaluated is \textbf{Reka Core} itself. This example show that the evaluator needs to be fairly capable in reasoning, being able to assign partial credit for measuring the scissors using the measuring tape.}
\label{table:example_scissors}
\end{figure}

%% file: tables/vibe_score.tex
\begin{table}[ht]
\caption{Vibe-Eval score and ranking for existing multimodal language models sorted by overall score.\newline$^\dagger$Note we expect the results of Reka Core to be worse on the \texttt{hard}-set, as these are, by their very definition, prompts that Core cannot solve.}
\centering
\setlength{\tabcolsep}{10pt} %
\rowcolors{3}{lightgray}{white}  %

\begin{tabular}{@{}lclc@{}}
\toprule
\multicolumn{1}{c}{\multirow{2}{*}{Model}} & \multicolumn{3}{c}{Vibe-Eval Score (\%)}        \\
\multicolumn{1}{c}{}                       & \texttt{all} & \texttt{hard} & \texttt{normal} \\ \midrule
Gemini Pro 1.5 & $60.40$ & $53.00$ & $64.80$ \\
GPT-4V & $57.90$ & $46.00$ & $64.90$ \\
Reka Core & $53.70$ & $38.20^\dagger$ & $62.80$ \\
Claude Opus & $52.80$ & $41.80$ & $59.20$ \\
Reka Flash & $52.20$ & $39.20$ & $59.90$ \\
Claude Sonnet & $52.10$ & $39.70$ & $59.50$ \\
Claude Haiku & $49.80$ & $38.50$ & $56.40$ \\
Llava-1.6-34b & $48.60$ & $39.90$ & $53.70$ \\
Reka Edge & $45.40$ & $32.20$ & $53.10$ \\
Llava-1.6-7b & $43.70$ & $35.30$ & $48.60$ \\
Idefics-2-8b & $40.00$ & $32.20$ & $44.60$ \\
Idefics-1-80b & $36.00$ & $32.10$ & $38.30$ \\
Fuyu-8b & $30.80$ & $23.40$ & $35.20$ \\ 
\bottomrule
\end{tabular}

\label{vibeeval}
\end{table}

%% file: tables/human_rankings.tex
\begin{table}[H]
\setlength{\tabcolsep}{10pt} %
\rowcolors{4}{lightgray}{white}  %
\caption{Rankings for Human preference on Vibe-Eval, measured in ELO from pairwise preferences. ELO scores are computed on all prompts, as well as the disjoint ``hard" and ``regular" difficulty subsets.
}
\centering
\begin{tabular}{@{}lccc@{}}
\toprule
\multirow{2}{*}{Model} & \multicolumn{3}{c}{Human Preference (ELO)}                      \\ \cmidrule(l){2-4} 
                       & \texttt{all} & \texttt{hard} & \texttt{normal} \\ \midrule
GPT-4V & $1290$  & $1224$ & $1277$ \\
Gemini Pro 1.5 & $1254$ & $1207$ & $1356$\\
Reka Flash & $1133$ & $1080$ & $1163$ \\
Claude 3 Opus & $1125$ & $1134$ & $1117$ \\
Reka Core & $1109$ & $1009$ & $1168$ \\
Claude 3 Sonnet & $1060$ & $1051$ & $1060$ \\
Claude 3 Haiku  & $1022$ & $1000$ & $1034$ \\
Llava 1.6 34B & $993$ & $1025$ & $974$ \\
Reka Edge & $990$ & $987$ &   $988$ \\ 
Llava 1.6 7B & $924$ & $906$ & $931$ \\ 
Idefics-2 8B & $774$ & $862$ & $705$\\ 
Idefics-1 80b & $761$ & $857$ & $705$\\
Fuyu-8b & $566$ & $660$ & $509$\\ 
\bottomrule
\end{tabular}
\label{humanrankings}
\end{table}

%% file: tables/correlation.tex
\begin{table}[H]
\setlength{\tabcolsep}{10pt} %
\rowcolors{4}{lightgray}{white}  %
\caption{Mappings of Human Preference rankings to VibeEval ranks. Our experiments show that Human preference ranking of models roughly correlate to VibeEval rankings.}
\centering
\begin{tabular}{@{}lccc@{}}
\toprule
\multirow{2}{*}{Model} & \multicolumn{3}{c}{Human Preference rank $\rightarrow$ VibeEval rank}                      \\ \cmidrule(l){2-4} 
                        & \texttt{hard} & \texttt{normal} \\ \midrule
GPT-4V & $[1-2] \rightarrow [1-4]$ &   $[1] \rightarrow [1-4]$ \\
Gemini Pro 1.5 & $[1-2] \rightarrow [1-2]$ &  $[2] \rightarrow [1-4]$     \\
Reka Flash & $[3-6] \rightarrow [2-7]$ & $[3-5] \rightarrow [2-7]$ \\
Claude 3 Opus & $[3-4] \rightarrow [2-7]$ &  $[4-5] \rightarrow [2-7]$ \\
Reka Core & $[5-9] \rightarrow [3-9]$ & $[3-5] \rightarrow [1-6]$ \\
Claude 3 Sonnet & $[4-8] \rightarrow [2-7]$ &  $[6-7] \rightarrow [2-7]$ \\
Claude 3 Haiku &  $[6-9] \rightarrow [4-9]$ &   $[6-8] \rightarrow [4-9]$ \\
Llava 1.6 34B & $[4-9] \rightarrow [2-9]$ &  $[8-10] \rightarrow [6-10]$ \\
Reka Edge & $[7-9] \rightarrow [7-10]$ & $[7-9] \rightarrow [7-10]$ \\
Llava 1.6 7B & $[10-12] \rightarrow [9-11]$   & $[9-10] \rightarrow [9-11]$  \\ 
Idefics-2 8B & $[10-12] \rightarrow [10-11]$   & $[11-12] \rightarrow [12]$\\
Idefics-1 80b  & $[10-12] \rightarrow [12-13]$   & $[11-12] \rightarrow [11]$\\
Fuyu-8b  & $[13] \rightarrow [13]$   & $[12-13] \rightarrow [13]$\\ \bottomrule
\end{tabular}
\label{rankingmappings}
\end{table}

\begin{table}[htbp]
    \newcommand{\agreement}[2]{#1 / #2}

    \caption{
        Agreement rate between Reka Core as a judge (Automatic), human annotators, and expert human annotators, on normal and hard subsets (given as \textit{normal} / \textit{hard}). The agreement rate is the percentage of times that two techniques share the same preference for model A vs model B over all prompts, skipping cases where at least one technique predicts a tie. A preference stronger than the weakest possible is required to not count as a tie.
        \label{table:agreement}
    }
    \centering
    \begin{tabular}{rccc}
         \toprule
        & Automatic & Human & Human Expert \\ \cmidrule{2-4}
        Automatic & \agreement{100.0\%}{100.0\%} & \agreement{94.2\%}{97.2\%} & \agreement{-}{99.8\%} \\

        Human &  & \agreement{99.3\%}{99.1\%} & \agreement{-}{96.6\%} \\

        Human Expert & &  & \agreement{-}{99.9\%} \\ \midrule
        Tie rate & \agreement{77.7\%}{80.9\%} &  \agreement{71.3\%}{76.7\%} & \agreement{-}{74.2\%} \\ \bottomrule
    \end{tabular}

\end{table}

%% file: tables/prompt_ambiguity.tex
\begin{figure}[h!]
\centering
\begin{tabular}[]{p{0.7in}p{5in}}
\toprule
\multicolumn{2}{p{6in}}{
\textbf{Example of hard prompt in Vibe-Eval}:}\\
\midrule
Image &
\includegraphics[width=0.75\linewidth]{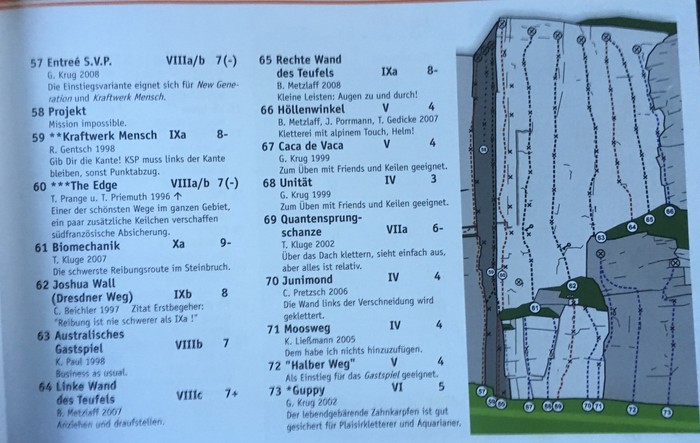} \vspace{0.5em}\\
Text prompt &
I want to climb the route that reaches the highest point, what should I climb? Give the name and required grade \vspace{0.5em}\\

Reference Response & The line that reaches the highest vertical point is climb \#60, named "The Edge" with a grade of VIIIa/b or 7(-). \vspace{0.5em}\\ 

Reka Core & The route that reaches the highest point according to the image is ``Rechte Wand des Teufels'' (The Devil's Right Wall). It is rated IXa, 8-. This route is described as having small ledges, requiring the climber to close their eyes and push through, and is suitable for practicing with cams and nuts. \vspace{0.5em} \\

GPT-4V & To reach the highest point on the climbing wall shown in the image, you should climb the route named \"Biomechanik\" which is numbered as 61. This route is graded Xa, indicating a very high level of difficulty. It is described as the hardest friction route in the area, so it requires advanced climbing skills. \vspace{0.5em} \\

Gemini\newline Pro 1.5 & Kraftwerk Mensch, IXa \vspace{0.5em} \\

Claude Opus & Based on the information provided in the climbing route list, the route that reaches the highest grade is ``Kraftwerk Mensch'' graded IXa (8-) by R. Gentsch in 1998. The description mentions ``Gib Dir die Kante! KSP muss links der Kante bleiben, sonst Punktabzug.'', which roughly translates to ``Give yourself the edge! KSP must stay left of the edge, otherwise point deduction.''\newline So if you want to climb the most challenging route that reaches the highest grade on this list, ``Kraftwerk Mensch'' graded IXa (8-) is the one to attempt.\\

\bottomrule

\end{tabular}
\caption{Example of a \texttt{hard}-set prompt where all models answer incorrectly. All models fail to reason that text on the left side of the image is a legend for the diagram (``toppo'') on the right. Image credit:~\cite{krug2014rotgelbes}, a climbing guidebook.}
\label{table:example_climbing}
\end{figure}

%% file: tables/prompt_inverse_scaling2.tex
\begin{figure}[!h]
\centering
\begin{tabular}[]{p{0.7in}p{5in}}
\toprule
\multicolumn{2}{p{6in}}{
\textbf{Example I of inverse scaling in Vibe-Eval}:}\\
\midrule
Image &
\includegraphics[width=0.5\linewidth]{images/bear_hat.jpg} \newline\\
Text prompt & what colour is my hat (im sitting on the bear)\newline\newline \\

Reference Response & The hat you are wearing is blue.
\newline\newline\\ 

Reka Core & You are wearing a black hat while sitting on the large teddy bear in the image.\newline\\

GPT-4V & Your hat is black. \newline\\

Claude Opus &  The image shows a large brown teddy bear statue or sculpture inside what appears to be a store, likely a toy or gift shop based on the shelves full of plush toys and other merchandise in the background. On top of the bear's head sits a small yellow bird plush toy wearing a red hat. There are no humans visible in the image that I can identify.\newline \\

Reka Edge &  The hat you are sitting on the bear is blue.\newline\\

Idefics 2 & Blue. \\

\bottomrule

\end{tabular}
\caption{Example of a difficult prompt that exhibits inverse scaling, i.e. all large frontier models (GPT4-V, Gemini Pro 1.5 and Reka Core) elicit incorrect responses, in contrast, Reka Edge and Idefics-2, small 7B and 8B parameter models respectively, both correctly identify the colour of the hat.}
\label{table:example_bear_hat}
\end{figure}

%% file: tables/prompt_inverse_scaling.tex
\begin{figure}[p]
\centering
\begin{tabular}[]{p{0.7in}p{5in}}
\toprule
\multicolumn{2}{p{6in}}{
\textbf{Example II of inverse scaling in Vibe-Eval}:}\\
\midrule
Image &
\includegraphics[width=0.5\linewidth]{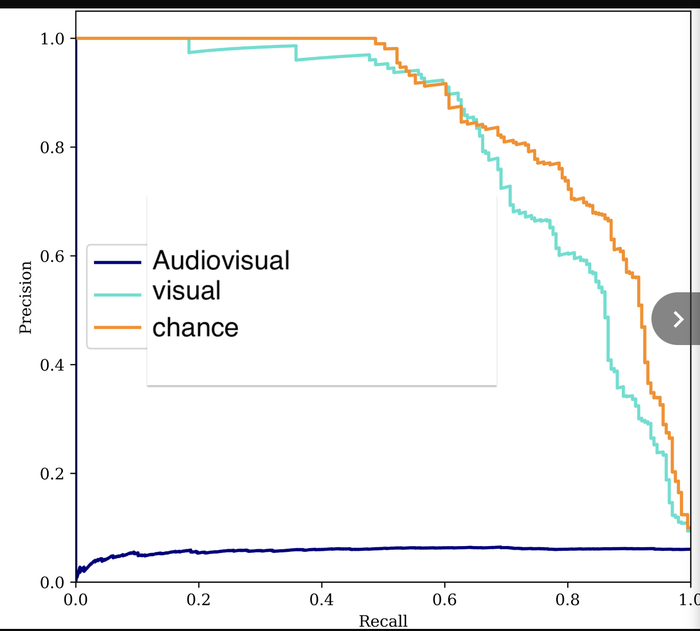} \vspace{0.5em}\\
Text prompt & rank the methods from best to worst\newline\vspace{0.5em} \\

Reference Response & The methods can be ranked from best to worst based on their performance area under the Precision-Recall curve, which is a common evaluation metric for classification tasks, especially in the context of imbalanced datasets. The larger the area under the curve, the better the performance of the method.\newline
1. chance: The chance method has the largest area under the curve.

2. visual: The audio method has slightly less area under the curve than "chance".

3. Audiovisual: This method has low precision across all levels of recall, indicating that it performs significantly worse than the "chance" and "visual" methods. In fact the constant low precision indicates the model is not working at all.\newline\\ 

Reka Core & The methods can be ranked from best to worst based on the area under the Precision-Recall curve (AUC-PR). The higher the AUC-PR...\textcolor{mydarkblue}{<truncated>}\newline
1. \textbf{Audiovisual:} This method has the highest AUC-PR...\textcolor{mydarkblue}{<truncated>}\newline
2. \textbf{Visual:} The visual method also performs well, with a high AUC-PR...\textcolor{mydarkblue}{<truncated>}\newline
3. \textbf{Chance:} The chance method has the lowest AUC-PR...\textcolor{mydarkblue}{<truncated>}.\newline
\\

GPT-4V & Based on the precision-recall curve in the image, the ranking of the methods from best to worst is as follows:\newline
1. \textbf{Audiovisual (blue line)} - This method consistently maintains...\textcolor{mydarkblue}{<truncated>}.\newline
2. \textbf{Visual (teal line)} - This method performs better than chance...\textcolor{mydarkblue}{<truncated>}\newline
3. \textbf{Chance (orange line)} - This method shows the lowest performance...\textcolor{mydarkblue}{<truncated>}\newline
\\

Gemini\newline Pro 1.5 & \#\# Ranking the Methods Based on the Precision-Recall Curve:\newline
Here's how we can rank the methods (Audiovisual, Visual, Chance)...\textcolor{mydarkblue}{<truncated>}
\newline
\textbf{1. Audiovisual:} This method clearly outperforms the others...\textcolor{mydarkblue}{<truncated>}\newline
\textbf{2. Visual:} This method performs decently but is inferior to...\textcolor{mydarkblue}{<truncated>}\newline 
\textbf{3. Chance:} This represents a baseline with no predictive power...\textcolor{mydarkblue}{<truncated>}
Therefore, the ranking from best to worst is: \textbf{Audiovisual > Visual > Chance} \newline \\

Reka Edge &  The chance method is the best, followed by the visual method, and then the audio  visual method is the worst.\newline\\

Idefics-2 & Chance, visual, audiovisual. \\

\bottomrule

\end{tabular}
\caption{Example of a difficult prompt that exhibits inverse scaling, i.e. all large frontier models (GPT4-V, Gemini Pro 1.5 and Reka Core) elicit incorrect responses, in contrast, Reka Edge and Idefics-2, small 7B and 8B parameter models respectively, both correctly rank the methods.}
\label{table:example_pr_curve}
\end{figure}

%% file: tables/prompt_partial_credit3.tex
\begin{figure}[!h]
\centering
\begin{tabular}[]{p{0.7in}p{5in}}
\toprule
\multicolumn{2}{p{6in}}{
\textbf{Example I of awarding partial credit in Vibe-Eval}:}\\
\midrule
Image &
\includegraphics[width=0.5\linewidth]{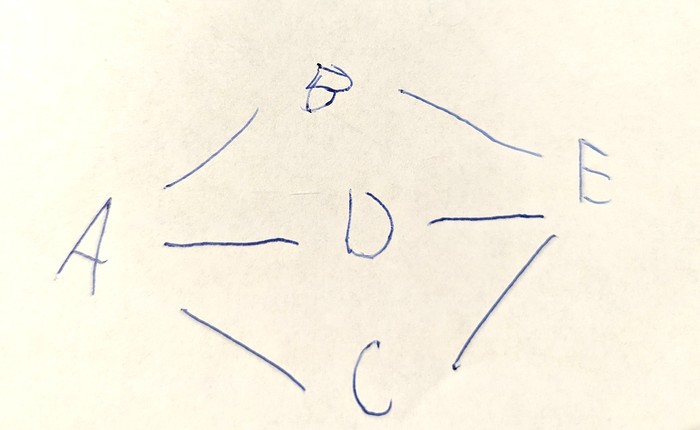} \vspace{0.5em}\\
Prompt & Is it possible to find Euler path in this graph? If yes, tell me node by node\newline \\

Reference response & An Euler path in a graph is a path that uses every edge exactly once. For a connected graph to have an Euler path, it must have exactly zero or two vertices of odd degree (number of edges incident to the vertex).\newline
Nodes A and E have odd degree (3), so possible path is:\newline
A -> B -> E -> D -> A -> C -> E\newline\\ 

Reka Core &  An Euler path in a graph is a path that uses every edge exactly once. A graph has an Euler path if and only if it is connected and has exactly 0 or 2 vertices of odd degree (number of edges incident to the vertex).\newline
In the given graph, we have the following degrees for each vertex:\newline
- A: 3 (odd)\newline
- B: 2 (even)\newline
- C: 3 (odd)\newline
- D: 2 (even)\newline
- E: 1 (odd)\newline
Since there are more than 2 vertices with an odd degree (A, C, E), this graph does not have an Euler path. \newline\\

Evaluator & Explanation: The assistant response initially correctly explains the conditions for a graph to have an Euler path. However, the assistant's conclusion that the graph does not have an Euler path is incorrect. The ground truth response correctly identifies the Euler path in the graph: A -> B -> E -> D -> A -> C -> E.\newline
Rating: \textbf{3}\newline
The assistant response is partially correct in explaining the conditions for an Euler path but fails to identify the correct Euler path in the given graph. The ground truth response is correct, but the assistant response could have been more accurate and complete.\newline\\

\bottomrule

\end{tabular}
\caption{Example of evaluator giving partial credit for model response.}
\label{table:euler_example}
\end{figure}

%% file: tables/prompt_partial_credit.tex
\begin{figure}[!h]
\centering
\begin{tabular}[]{p{0.7in}p{5in}}
\toprule
\multicolumn{2}{p{6in}}{
\textbf{Example II of awarding partial credit in Vibe-Eval}:}\\
\midrule
Image &
\includegraphics[width=0.5\linewidth]{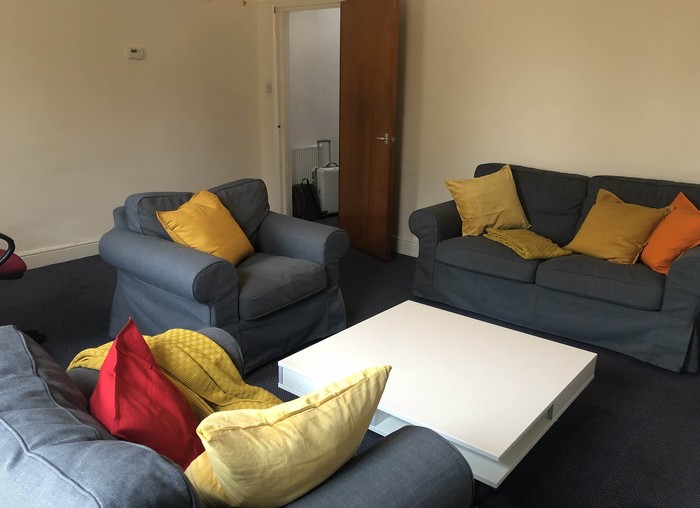} \vspace{0.5em}\\
Text prompt & List the number of pillows for each colour in the rainbow\newline\newline \\

Reference Response & In the image, there are:\newline- Yellow pillows: 4\newline- Orange pillows: 1\newline- Red pillows: 1\newline\\ 

Reka Core &  There are 3 yellow pillows, 1 red pillow, and 1 orange pillow. \newline\\

Evaluator & Explanation: The assistant response is partially correct but has some inaccuracies. The correct number of pillows for each color in the rainbow, based on the ground truth response, is:\newline
Yellow pillows: 4\newline
Orange pillows: 1\newline
Red pillows: 1\newline The assistant response missed the correct number of yellow pillows and did not mention that there are no pillows in the other colors of the rainbow (green, blue, indigo, violet) visible in the image.\newline
Rating: \textbf{3}\newline
\\
\bottomrule

\end{tabular}
\caption{Example of the evaluator giving partial credit for a model response.}
\label{table:example_pillows}
\end{figure}